\documentclass[runningheads]{llncs}
\usepackage{graphicx}
\usepackage{amsmath,amssymb} % define this before the line numbering.
\usepackage{color}
\usepackage{multirow}

\begin{document}
\pagestyle{headings}
\mainmatter

\def\ACCV20SubNumber{188}  % Insert your submission number here

%===========================================================
\title{Dense Dual-Path Network for Real-time Semantic Segmentation} % Replace with your title
\titlerunning{Dense Dual-Path Network for Real-time Semantic Segmentation}
\author{Xinneng Yang\inst{1} \and
Yan Wu\inst{*,1} \and
Junqiao Zhao\inst{1,2} \and
Feilin Liu\inst{1}}
\authorrunning{X. Yang et al.}
\institute{College of Electronics and Information Engineering, Tongji University, Shanghai, China \email{yanwu@tongji.edu.cn} \and
Institute of Intelligent Vehicle, Tongji University, Shanghai, China}

\maketitle

%===========================================================
\begin{abstract}
Semantic segmentation has achieved remarkable results with high computational cost and a large number of parameters. 
However, real-world applications require efficient inference speed on embedded devices. 
Most previous works address the challenge by reducing depth, width and layer capacity of network, which leads to poor performance. 
In this paper, we introduce a novel Dense Dual-Path Network (DDPNet) for real-time semantic segmentation under resource constraints. 
We design a light-weight and powerful backbone with dense connectivity to facilitate feature reuse throughout the whole network and the proposed Dual-Path module (DPM) to sufficiently aggregate multi-scale contexts. 
Meanwhile, a simple and effective framework is built with a skip architecture utilizing the high-resolution feature maps to refine the segmentation output and an upsampling module leveraging context information from the feature maps to refine the heatmaps. 
The proposed DDPNet shows an obvious advantage in balancing accuracy and speed. 
Specifically, on Cityscapes test dataset, DDPNet achieves 75.3\% mIoU with 52.6 FPS for an input of \(1024\times2048\) resolution on a single GTX 1080Ti card. 
Compared with other state-of-the-art methods, DDPNet achieves a significant better accuracy with a comparable speed and fewer parameters.
\end{abstract}

%===========================================================
\section{Introduction}

Semantic segmentation is a fundamental task in computer vision, the purpose of which is to assign semantic labels to each image pixel.
It has many potential applications in the fields of autonomous driving, video surveillance, robot sensing and so on.
Existing methods mainly focus on improving accuracy. 
However, these real-world applications require efficient inference speed on high-resolution images. 

Previous works\cite{chen2018encoder,cheng2019spgnet,fu2019dual,he2019dynamic,tian2019decoders,zhao2017pyramid} have already obtained outstanding performances on various benchmarks\cite{brostow2009semantic,cordts2016cityscapes,everingham2010pascal,lin2014microsoft,zhou2017scene}.
By analyzing existing state-of-the-art semantic segmentation methods, we find the keys to achieving high accuracy. 
1) The backbone of these methods has a powerful feature extraction capability, such as ResNet\cite{he2016deep}, ResNeXt\cite{xie2017aggregated}, DenseNet\cite{huang2017densely}, which is usually pre-trained on ImageNet\cite{deng2009imagenet}. 
These backbones have a strong generalization capability and thus are adapted to many computer vision tasks. 
2) These methods aggregate multi-scale context information sufficiently. 
There are many objects in semantic segmentation that are difficult to be distinguished only by their appearance, such as `field' and `grass', `building' and `wall'. 
Due to multiple scales, occlusion and illumination, some objects like `car' and `person' require multi-scale context information to be correctly identified. 
To address above issues, dilated convolution and pooling operation are often used to enlarge the receptive field. 
Even though the theoretical receptive field is large enough, it still can't fully exploit the capability of global context information. 
Therefore, some approaches\cite{chen2018encoder,he2019dynamic,zhao2017pyramid,chen2017deeplab,chen2017rethinking} aggregate multi-scale contexts via fusing feature maps generated by parallel dilated convolutions and pooling operations to robustly segment objects at multiple scales.
3) These methods recover spatial information effectively. 
Downsampling enlarges the receptive field and decreases the size of feature maps. 
It enriches high-level features, but loses spatial details. 
However, detailed spatial information is essential for semantic segmentation. 
In order to preserve spatial information, most works\cite{chen2018encoder,fu2019dual,he2019dynamic,zhao2017pyramid} remove the last two pooling operations and replace the subsequent convolutions with dilated convolutions to keep the receptive field unchanged at the expense of computational efficiency. 
Unlike them, \cite{cheng2019spgnet,tian2019decoders,fu2019adaptive,lin2017refinenet,long2015fully,noh2015learning,wang2019carafe} utilize upsampling methods and self-designed skip connection to refine the boundaries of objects and small objects.

Based on the above analysis, we summarize the keys to achieving high accuracy in semantic segmentation as follows:
\begin{itemize}
\item Backbone with a powerful feature extraction capability.
\item Sufficient aggregation of context information.
\item Effective recovery of spatial information.
\end{itemize}

Recently, real-time semantic segmentation methods\cite{li2019dabnet,li2019dfanet,liu2019feature,orsic2019defense,yu2018bisenet,zhuang2019shelfnet} have shown prom-ising results. 
\cite{wu2017real,zhao2018icnet} reduce the input size to accelerate the model, while easily losing the spatial details around boundaries and small objects.
\cite{mehta2018espnet,paszke2016enet} remove some of downsampling operations to create an extremely small model. 
Nevertheless, the receptive field of these models is not sufficient to cover large objects, resulting in a poor performance. 
To achieve real-time speed, some works\cite{li2019dabnet,liu2019feature,mehta2018espnet,paszke2016enet,romera2017erfnet} design a specialized network for semantic segmentation as backbone.
Differently, some works\cite{li2019dfanet,orsic2019defense,yu2018bisenet,zhuang2019shelfnet,siam2018comparative} adopt a light-weight classification network as backbone, such as MobileNets\cite{howard2019searching,howard2017mobilenets,sandler2018mobilenetv2}, ShuffleNets\cite{ma2018shufflenet,zhang2018shufflenet}, ResNet-18\cite{he2016deep}.
Convolution factorization refers to the decomposition of a large convolution operation into several smaller operations, such as factorized convolution and depth-wise separable convolution, which is widely adopted in these backbones to reduce the computational cost and the number of parameters. 
However, convolution factorization is not conducive to GPU parallel processing, which results in a much slower inference speed under the same computational budget.
On the other hand, these backbones have a limited capability due to fewer convolution operations and feature maps. 
Recent works\cite{yu2018bisenet,zhao2018icnet,poudel2019fast} propose a two-column network which consists of a deep network for encoding context information and a shallow network for capturing spatial information. 
However, the extra network on high-resolution images limits the inference speed, and the independence between networks limits the performance of the model.

To strike a better balance between accuracy and efficiency, we follow the principle of simplicity in designing the model. 
A complicated and sophisticated work may improve accuracy, but in most cases it hurts efficiency significantly. 
A simple and clean framework can make it easier to re-implement and improve. 
Therefore, we propose a light-weight yet powerful backbone and a simple yet effective framework for fast and accurate semantic segmentation. 
The proposed Dense Dual-Path Network (DDPNet) achieves 75.3\% mIoU with merely 2.53 M parameters on Cityscapes test dataset. 
It can run on high-resolution images (\(1024\times2048\)) at 52.6 FPS on a single GTX 1080Ti card. 
DDPNet is superior to most of the state-of-the-art real-time semantic segmentation methods in accuracy and speed, and requires fewer parameters.

Our main contributions are summarized as follows:
\begin{itemize}
\item We design a light-weight and powerful backbone with dense connectivity to facilitate feature reuse throughout the whole network and the proposed Dual-Path module (DPM) to sufficiently aggregate multi-scale contexts.
\item We propose a simple and effective framework with a skip architecture utilizing the high-resolution feature maps to refine the segmentation output and an upsampling module leveraging context information from the feature maps to refine the heatmaps.
\item We conduct a series of experiments on two standard benchmarks, Cityscapes and CamVid, to investigate the effectiveness of each component of our proposed DDPNet and compare accuracy and efficiency with other state-of-the-art methods.
\end{itemize}

%===========================================================
\section{Related Work}

Recently, FCN\cite{long2015fully} based methods have greatly improved the performance of semantic segmentation. 
Most of them focus on encoding content information and recovering spatial information.

\textbf{Real-time Segmentation:} 
The goal of real-time semantic segmentation is to achieve the best trade off between accuracy and efficiency. 
In order to reach a real-time speed, SegNet\cite{badrinarayanan2017segnet} and U-Net\cite{ronneberger2015u} perform multiple downsampling operations to significantly reduce the feature map size. 
SegNet designs a symmetric encoder-decoder architecture to carefully recover feature maps. 
U-Net proposes a symmetric architecture with skip connection to enable precise localization. 
Differently, E-Net\cite{paszke2016enet} constructs an extremely light-weight network with fewer downsampling operations to boost the inference speed. 
ERFNet\cite{romera2017erfnet} focuses on a better accuracy with a deeper network that uses residual connection and factorized convolution. 
ESPNet\cite{mehta2018espnet} proposes a light-weight network with efficient spatial pyramid module. 
ICNet\cite{zhao2018icnet} uses an image cascade network to capture objects of different sizes from multi-scale images. 
BiSeNet\cite{yu2018bisenet} designs a spatial path to preserve spatial information and a context path to obtain sufficient receptive field. 
Based on multi-scale feature propagation, DFANet\cite{li2019dfanet} reduces the number of parameters and maintains high accuracy.

\textbf{Context Information:} 
Semantic segmentation needs to sufficiently obtain context information to correctly identify objects that are similar in appearance but belong to different categories and objects that are different in appearance but belong to the same category. 
Most works capture diverse context information by using different dilation convolutions to enlarge the receptive field. 
DeepLab\cite{chen2017deeplab} proposes an atrous spatial pyramid pooling module to aggregate multi-scale contexts. 
In the follow-up work, \cite{chen2017rethinking} further improves performance by extending the previously proposed atrous spatial pyramid pooling module with a global average pool, which is able to capture the global context of images. 
Similarly, PSPNet\cite{zhao2017pyramid} proposes a pyramid pooling module which consists of different sizes of average pooling operations to aggregate multi-scale contexts. 
\cite{zhang2017scale} designs a scale-adaptive convolution to acquire flexible-size receptive fields. 
PAN\cite{li2018pyramid} combines attention mechanism and spatial pyramid to learn a better feature representation. 
DMNet\cite{he2019dynamic} adaptively captures multi-scale contents with multiple dynamic convolutional modules arranged in parallel.

\textbf{Spatial Information:} 
Semantic segmentation requires spatial details to restore the boundaries of objects and small objects. 
The reason for the loss of spatial details is downsampling operations in the convolutional network. 
Downsampling is essential for convolutional networks because it can reduce the size of feature maps and enlarge the receptive field. 
Most works reduce the number of downsampling operations to preserve spatial details, which leads to slow inference speed. 
Differently, \cite{noh2015learning,paszke2016enet,romera2017erfnet,badrinarayanan2017segnet} construct an encoder-decoder structure with unpooling and deconvolution to recover spatial details. 
However, this structure still can not effectively recover the loss of spatial details and have high computational complexity. 
Skip connection is first introduced in FCN\cite{long2015fully}, which combines semantic information from a deep layer with appearance information from a shallow layer to produce accurate and detailed segmentation result. 
Based on FCNs, RefineNet\cite{lin2017refinenet} presents a multi-path refinement network that refines high-level semantic features using long-range residual connection. 
BiseNet\cite{yu2018bisenet} and Fast-SCNN\cite{poudel2019fast} explicitly acquire spatial information at a fast inference speed using a light-weight spatial path.

%===========================================================
\section{Dense Dual-Path Network}

In this section, we introduce the backbone of DDPNet and the proposed framework for real-time semantic segmentation.
Furthermore, we elaborate the design choices and motivations in detail.

%------------------------------------------------------------------------- 
\subsection{Dense Connectivity}

The backbone of DDPNet is a variant of densely connected convolutional network, which adopts dense connectivity to stack convolution operations. 
Dense connectivity is originally proposed by DenseNet\cite{huang2017densely}, in which each layer has direct connections to all subsequent layers. 
Formally, the \(m^{th}\) layer takes feature maps from all preceding layers as input:
\begin{equation}
    x_m = H_m([x_0,x_1,...,x_{m-1}]) 
\end{equation}
where \(x_n\), \(n \in[1,2,...,m]\) refers to the output of \(n^{th}\) layer. 
\(x_0\) refers to the input of dense block, bracket indicates concatenation operation, and \(H\) is a composite function of three consecutive operations: batch normalization\cite{ioffe2015batch}, followed by a rectified linear unit\cite{glorot2011deep} and a convolution operation.

Some works\cite{jegou2017one,yang2018denseaspp} utilize dense connectivity in the network to boost the performance of semantic segmentation. 
However, they focus on accuracy rather than speed.  
Dense connectivity allows for the reuse of features throughout the network and significantly reduces the number of parameters, which contributes to the implementation of light-weight models. 
Therefore, we use dense connectivity to build a light-weight and  powerful backbone for real-time semantic segmentation. 
Following PeleeNet\cite{wang2018pelee}, we use post-activation as composite function, which conducts convolution operation before batch normalization, for improving inference speed.

%------------------------------------------------------------------------- 
\subsection{Dual-Path Module}

\begin{figure}[t]
\centering
\includegraphics[height=5.0cm]{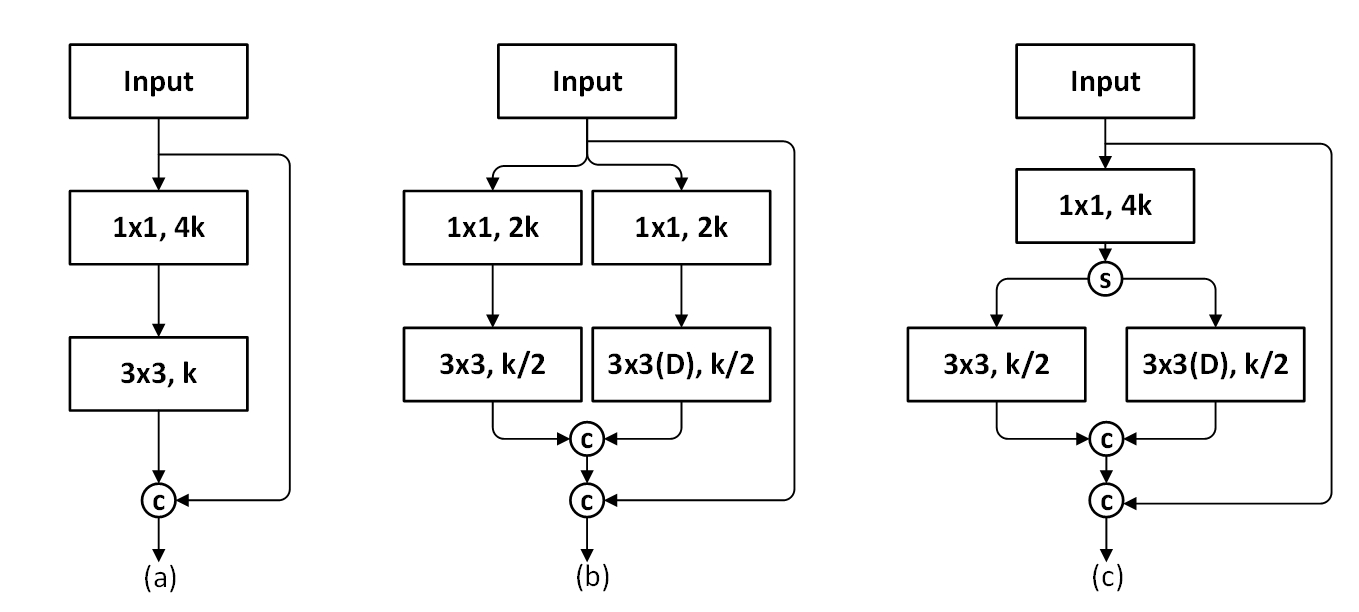}
\caption{ 
Depiction of the dense layer originally proposed in DenseNet\cite{huang2017densely}, and our proposed Dual-Path module (DPM). 
``\(k\)'': the growth rate. 
``\(D\)'': dilated convolution. 
``\(c\)'': channel concatenate operation. 
``\(s\)'': channel split operation.
In the convolutional blocks, ``\(s1 \times s2, n_o\)'' indicates their kernel sizes \(\left(s_1, s_2\right)\) and the number of output feature maps \(\left(n_o\right)\). 
(a) Original dense layer. 
(b-c) Dual-Path Module. 
For brevity, we omit normalization and activation function.
}
\label{fig:fig1}
\end{figure}

Dual-Path module (DPM) is the basic unit of dense block. 
Motivated by the diversity of object scales in semantic segmentation, we propose a specific Dual-Path module composed of two bottleneck layers in parallel. 
One of the bottleneck layers uses a point-wise convolution to reduce the number of input feature maps, followed by a \(3\times3\) convolution to produce \(\frac{k}{2}\) new feature maps. 
\(k\) refers to the growth rate. 
The other bottleneck layer replaces the \(3\times3\) convolution with dilation convolutions to learn visual patterns for large objects.
The structural details are shown in Fig.~\ref{fig:fig1}(b).
We refer this structure as DPM-b. 
Notice that the point-wise convolutions in both branches have the same input. 
Therefore, we combine the two point-wise convolutions into one, and split the output of the convolution into two independent inputs for the two branches, as depicted in Fig.~\ref{fig:fig1}(c).
We refer this structure as DPM-c.
The two implementations are functionally identical but differ in efficiency.
DPM-c is more efficient than DPM-b in GPU parallel processing. 
For this reason, we adopt the structure of DPM-c as our final implementation of DPM.
Finally, the output of two branches is concatenated, followed by a dropout layer\cite{srivastava2014dropout}. 

As can be seen from Fig.~\ref{fig:fig1}, Dual-Path module has a larger effective receptive field than the original dense layer. 
With an extra dilated branch in dense layer, Dual-Path module can extract features from different scales. 
Intuitively, more branches can aggregate multi-scale contexts more effectively, such as ASPP in \cite{chen2017deeplab}. 
However, the decomposition of a single convolution into multiple parallel convolutions is not conducive to the acceleration of the model. 
Due to the ability to effectively aggregate feature maps at different scales, Dual-Path module significantly improves the capacity of backbone.

%------------------------------------------------------------------------- 
\subsection{Backbone Design}

In this subsection, we discuss the main components and algorithms used to build the backbone of Dense Dual-Path Network. 
In this work, we aim to design an architecture that gets the best possible trade-off between accuracy and efficiency. 
Many approaches that focus on designing a light-weight architecture largely adopt depth-wise separable convolution and factorized convolution which lack efficient implementation. 
Instead of using depth-wise separable convolution or factorized convolution, Dense Dual-Path Network is build with traditional convolution.

\textbf{Transition Layer.} 
Transition layer is used to reduce the size of feature maps and compress the model. 
It is composed of a point-wise convolution followed by a \(2\times2\) average pooling layer. 
In order to fully exploit dense connectivity, DDPNet keeps the number of input channels and output channels the same in all transition layers. 
This design facilitates feature reuse throughout the whole network.

\textbf{Initial Block.} 
Initial block is used to reduce the input size, which typically involves several downsampling operations. 
Due to direct operation on the original image, initial block is often computationally expensive. 
However, a well-designed initial block can effectively improve feature expression and preserve rich spatial information. 
The initial block of DPPNet is motivated by ENet\cite{paszke2016enet} and PeleeNet\cite{wang2018pelee}, which preserves rich spatial information and takes full advantage of feature reuse. 
In our initial block, a \(3\times3\) convolution with stride 2 is performed on the original image, followed by two branches. 
One of the branches is a \(3\times3\) convolution with stride 2. 
The other branch is a \(2\times2\) max pooling layer with stride 2. 
Finally, the output of two branches is concatenated.

\setlength{\tabcolsep}{4pt}
\begin{table}[t]
\begin{center}
\caption{
The backbone of our proposed DDPNet. 
``\(Dense Layer\left(or DPM\right) \times n_d, k\)'' indicates the operation in dense block is the original dense layer \(\left(Dense Layer\right)\) or the proposed Dual-Path module \(\left(DPM\right)\), the number of layers \(\left(n_d\right)\) and the growth rate \(\left(k\right)\). 
Input size is \(\left(1024\times2048\times3\right)\). 
}
\label{table:table1}
\begin{tabular}{l|c|r}
\hline\noalign{\smallskip}
\textbf{Type}                    & \textbf{Operator}            & \textbf{Output Shape}            \\
\noalign{\smallskip}\hline\noalign{\smallskip}
Initial Block                     &                              & 256 x 512 x 64                   \\ \noalign{\smallskip}\hline\noalign{\smallskip}
Dense Block                       & Dense Layer   x 2, k=32              & 256 x 512 x 128                  \\ \noalign{\smallskip}\hline\noalign{\smallskip}
\multirow{2}{*}{Transition Layer} & 1 x 1 conv, stride 1         & \multirow{2}{*}{128 x 256 x 128} \\ \cline{2-2}
                                  & 2 x 2 average pool, stride 2 &                                  \\ \noalign{\smallskip}\hline\noalign{\smallskip}
Dense Block                       & Dense Layer   x 4, k=32              & 128 x 256 x 256                  \\ \noalign{\smallskip}\hline\noalign{\smallskip}
\multirow{2}{*}{Transition Layer} & 1 x 1 conv, stride 1         & \multirow{2}{*}{64 x 128 x 256}  \\ \cline{2-2}
                                  & 2 x 2 average pool, stride 2 &                                  \\ \noalign{\smallskip}\hline\noalign{\smallskip}
Dense Block                       & DPM   x 8, k=32              & 64 x 128 x 512                   \\ \noalign{\smallskip}\hline\noalign{\smallskip}
\multirow{2}{*}{Transition Layer} & 1 x 1 conv, stride 1         & \multirow{2}{*}{32 x 64 x 512}   \\ \cline{2-2}
                                  & 2 x 2 average pool, stride 2 &                                  \\ \noalign{\smallskip}\hline\noalign{\smallskip}
Dense Block                       & DPM   x 8, k=32              & 32 x 64 x 768                    \\ \noalign{\smallskip}\hline\noalign{\smallskip}
Transition Layer                  & 1 x 1 conv, stride 1         & 32 x 64 x 768                    \\ \noalign{\smallskip}\hline
\end{tabular}
\end{center}
\end{table}
\setlength{\tabcolsep}{1.4pt}

\textbf{Architecture Detail.}
The backbone of our proposed DDPNet is shown in Table~\ref{table:table1}. 
The entire architecture consists of an initial block and four dense blocks followed by a transition layer. 
To maintain a better balance between accuracy and computational cost, DDPNet first reduces spatial resolution twice in the initial block and performs downsampling operation in each transition layer (except for the last one). 
Except for the last block, DDPNet doubles the number of feature maps in each dense block.

In DenseNet\cite{huang2017densely}, each layer produces \(k\) new feature maps, where \(k\) refers to as the growth rate. 
The growth rate is usually a small constant due to feature reuse. 
With a fixed number of output channels and a fixed growth rate, we can get the number of layers in a certain block. 
For example, the \(n^{th}\) block has \(\frac{(n_{out} - n_{in})}{k}\) layers, where \(n_{out}\) is the number of channels at the end of the block, and \(n_{in}\) is the number of channels in the input layer. 
As mentioned in \cite{huang2017densely,huang2018condensenet}, convolution layers in a deeper block tend to rely more on high-level features than on low-level features. 
Based on this observation, DDPNet produces more new feature maps in deeper blocks, which means that more layers are needed in deeper blocks. 

Since the feature maps from a higher layer contain more semantic information and less spatial information, we adopt the original dense layer in the first two blocks and replace the original dense layer with Dual-Path module in the last two blocks. 
We explore further in the experiment section.  
A larger growth rate requires fewer layers to generate new feature maps, which can boost the inference speed. 
However, a smaller growth rate forms a denser connections, which improves the quality of feature maps. 
To strike a better balance between accuracy and efficiency, we set the growth rate to 32 in DDPNet. 

%------------------------------------------------------------------------- 
\subsection{Framework for Real-time Semantic Segmentation}

\begin{figure}[t]
\centering
\includegraphics[height=8.5cm]{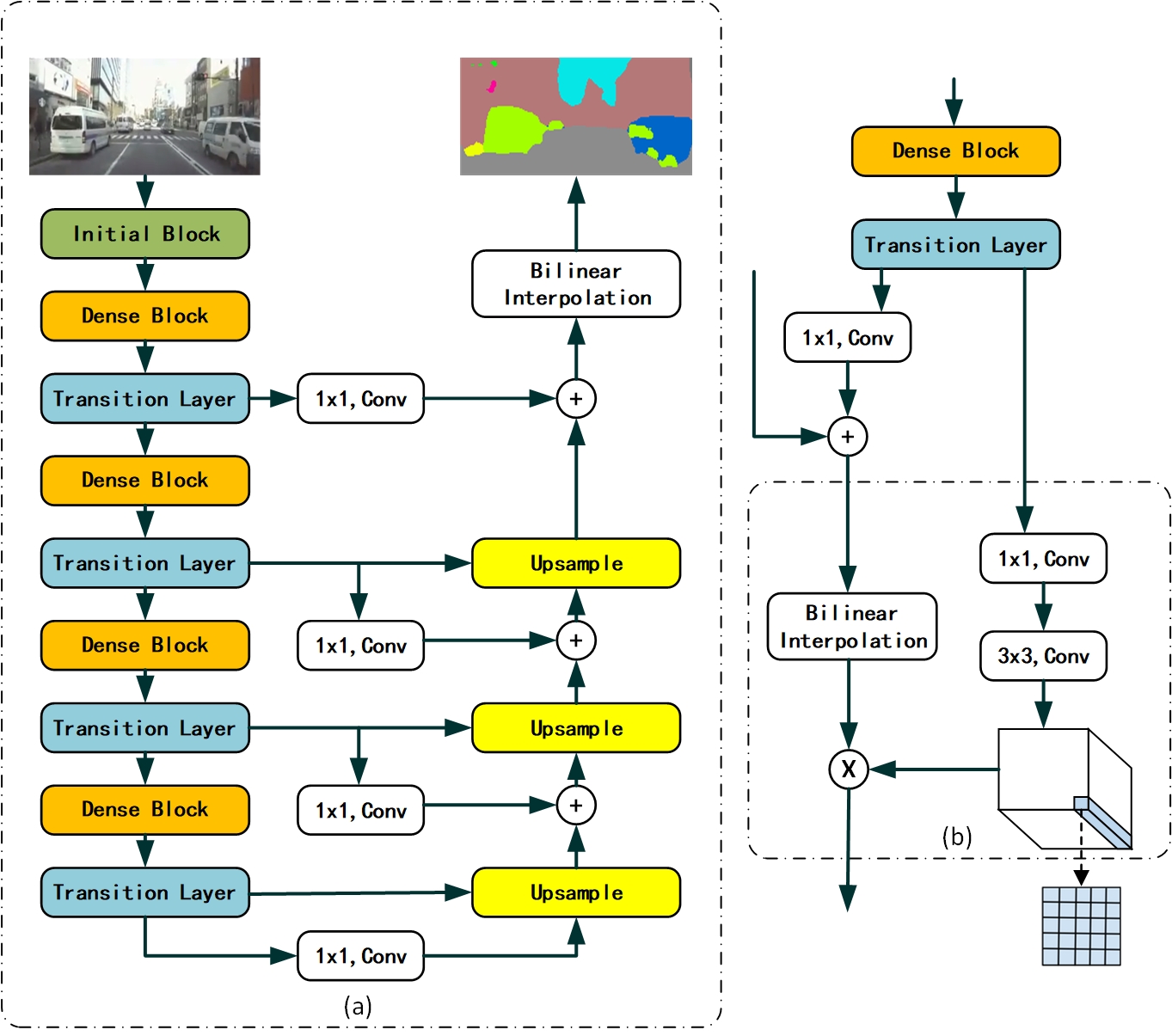}
\caption{ 
(a) Framework of the proposed DDPNet. 
(b) Architecture details of the usampling module. 
Note that the feature maps used to generate heatmaps are the output of the point-wise convolution in transition layer before downsampling. 
For brevity, we refer to these feature maps as the output of transition layer in this diagram. 
``\(+\)'': element-wise addition. 
``\(\times\)'': weighted sum operation.
}
\label{fig:fig3}
\end{figure}

In this subsection, we propose a simple and effective framework for real-time semantic segmentation. 
Fig.~\ref{fig:fig3}(a) shows the overall framework. 
The proposed backbone and the framework are adopted to construct the DDPNet. 
The backbone can be other classification networks, such as ResNet\cite{xie2017aggregated}. 
Most of the frameworks used for real-time semantic segmentation adopt an encoder-decoder architecture. 
The encoder is used to provide information for the final classification and reduce the computational cost by downsampling. 
The main purpose of the decoder is to recover the spatial details lost during the downsampling process.

\textbf{Skip Architecture.} 
DDPNet employs an asymmetric sequential architecture, where an extreme light-weight decoder is adopted to upsample the feature maps to match the input resolution. 
The decoder of DDPNet is a skip architecture, which utilizes the high-resolution feature maps to refine the segmentation output. 
In U-Net\cite{ronneberger2015u}, skip connection is performed on the feature space, which is very computational expensive because it is directly affected by the number of channels. 
To accelerate the inference speed, DDPNet adopts skip connection on heatmaps in the label space, similar to FCN\cite{long2015fully}. 
The output of each transition layer is followed by a point-wise convolution to map from feature space to label space, resulting in a heatmap corresponding to each class. 
Note that the feature maps used to generate heatmaps are the output of the point-wise convolution in transition layer before downsampling. 
For convenience, we refer to these feature maps as the output of transition layer in the following paper. 
A low-resolution heatmap is upsampled by an upsampling module, and then element-wise addition is performed between this heatmap and a high-resolution heatmap. 
After three skip connections, a bilinear interpolation with stride 4 is adopted to produce the final output. 
Fig.~\ref{fig:fig3}(a) shows the overall framework of DDPNet. 

\textbf{Upsampling Module.} 
The commonly used upsampling methods in semantic segmentation are bilinear interpolation and deconvolution. 
Bilinear interpolation is computationally cheap, but only leverages the distance information between pixels. 
Deconvolution uses fixed kernels to perform upsampling, which has been proved to be limited and inefficient. 
The upsampling module of DDPNet is modified from \cite{wang2019carafe} to save a lot of computation by performing upsampling on label space instead of upsampling on feature space. 
More specifically, DDPNet uses a bilinear interpolation to upsample the heatmap and refine the upsampled heatmap with a dynamic filtering layer. 
Dynamic filtering layer is originally proposed in dynamic filter networks\cite{jia2016dynamic}. 
In dynamic filter networks, a dynamic filter module consists of a filter generating path that produces filters based on one input, and a dynamic filtering layer that applies the generated filters to another input. 
In the upsampling module of DDPNet, the filter generating path takes the output of transition layer as input, then compresses the input feature channel with a point-wise convolution. 
Finally, a \(3\times3\) convolution followed by a pixel shuffle operation and a softmax is used to generate the filters. 
In dynamic filtering layer, the generated filters are applied to refine the upsampled heatmap, which is a weighted sum operation.
Formally, the upsampling module can be written as 
\begin{equation}
    \mathbf{Y} = U(\mathbf{X_1}) \otimes \delta(S(N(F^1(w_{3\times3},\sigma(N(F^2(w_{1\times1},\mathbf{X_2})))))))
\end{equation}
where \(Y\) and \(X\) are the output and input of the upsampling module respectively. 
\(X_1\) and \(X_2\) refer to the input of dynamic filtering layer and filter generating path. 
\(U\) denotes bilinear interpolation, \(\otimes\) indicates weighted sum operation, \(F\) and \(N\) represent convolution operation and batch normalization, \(S\) refers to pixel shuffle operation, \(\sigma\) and \(\delta\) indicate ReLU and softmax activation. 
\(w_{n\times n}\) is convolutional parameter and \(n\) represents kernel size. 
Fig.~\ref{fig:fig3}(b) shows the detailed architecture of the upsampling module.

%===========================================================
\section{Experiment} 

In this section, we evaluate the proposed DDPNet on Cityscapes\cite{cordts2016cityscapes} and CamVid \cite{brostow2009semantic} benchmarks. 
We first introduce the datasets and the implementation protocol. 
Then, we conduct a series of experiments on Cityscapes validation set to investigate the effectiveness of each component of our proposed method. 
Finally, we report the accuracy and efficiency results compared with other state-of-the-art methods. 
All accuracy results are reported using the commonly used mean Intersection-over-Union (IoU) metric. 
Runtime evaluations are performed on a single 1080Ti card.
To eliminate the error fluctuation, we report an average of 5000 frames for the frames per second (FPS) measurement.

\textbf{Cityscapes.} 
The Cityscapes dataset is a large urban street scene dataset which is collected from 50 different cities. 
It contains 5000 fine annotated images, which are split into three sets: 2975 images for training, 500 images for validation, and 1525 images for testing. 
For fair a comparison, the 1525 test images is offered without ground-truth. 
There is another set of 19,998 images with coarse annotation, but we only use the fine annotated images for all experiments. 
All images have a resolution of \(1024\times2048\), in which each pixel is annotated to pre-defined 19 classes. 

\textbf{CamVid.} 
The CamVid dataset is another street scene dataset which is extracted from video sequences. 
It contains 701 images, which are split into three sets: 367 images for training, 101 images for validation, and 233 images for testing. 
All images have a resolution of \(720\times960\) and 11 semantic categories. 
Following the general settings, we reduce the image resolution to \(360\times480\) and use the training set and the validation set to train our model, then test it on the test set.

%------------------------------------------------------------------------- 
\subsection{Implementation Protocol}

We conduct all experiments using PyTorch with CUDA 10.0 and cuDNN backends on a single 1080Ti card. 
We use Adam optimizer \cite{kingma2014adam} with batch size 8 and weight decay \(2e^{-4}\) in training. 
For learning rate schedule, we use the learning rate warmup strategy suggested by \cite{he2019bag} and the cosine learning rate decay policy. 
The learning rate \(l_{i}\) is computed as:
\begin{equation}
    l_{i} = \frac{1}{2} \times \left( 1 + \cos{ \frac{i \times \pi}{T}} \right) \times l_{base}
\end{equation}
where \(i\) refers to current epoch, the maximum number of epochs \(T\) is set to 350 for training on Cityscapes and 700 for training on CamVid, the initial learning rate \(l_{base}\) is \(5e^{-4}\). 
As for data augmentation, we employ mean subtraction, random horizontal flip and random scale on the input images during training. 
A random parameter between [0.75, 2] is used to transform the images to different scales. 
Finally, we randomly crop the images into fixed size for training.

%------------------------------------------------------------------------- 
\subsection{Ablation Study}

In this subsection, we perform a series of experiments to evaluate the effectiveness of each component in our proposed DDPNet. 
All ablation studies are trained on Cityscapes training set and evaluated on Cityscapes validation set. 
We reduce the image size to \(768\times1536\) to accelerate the training process and evaluate the accuracy and efficiency of our models. 
For a fair comparison, we use the same training settings for all models.

\setlength{\tabcolsep}{4.0pt}
\begin{table}[t]
\begin{center}
\caption{
Ablation studies for dense connectivity and DPM. 
All models are trained from scratch on Cityscapes training set and evaluated on Cityscapes validation set.
The numbers in brackets denote the performance improvement over the baseline. 
The bold entries are the final settings for DDPNet.
}
\label{table:table2}
\begin{tabular}{|l|c|c|c|c|c|c|}
\hline
Model                        & mIoU(\%)    & Time(ms) & FPS   & Para  & FLOPs & Memory \\
\hline
Baseline                     & 72.8        & 9.5      & 105.7 & 2.7M  & 13.7G & 333MB  \\
ResNet\cite{he2016deep}      & 70.9        & 9.8      & 101.6 & 11.2M & 41.7G & 287MB  \\
PeleeNet\cite{wang2018pelee} & 72.5        & 11.7     & 85.3  & 2.1M  & 12.3G & 406MB  \\
\hline
+ $DPM \times 1$             & 74.6 (+1.8) & 10.3     & 97.4  & 2.6M  & 13.5G & 333MB  \\
\textbf{+ DPM $\times$ 2 }   & 75.5 (+2.7) & 11.2     & 89.4  & 2.4M  & 12.8G & 333MB  \\
+ $DPM \times 4$             & 74.7 (+1.9) & 11.7     & 85.5  & 2.4M  & 11.5G & 333MB  \\
\hline
\end{tabular}
\end{center}
\end{table}
\setlength{\tabcolsep}{1.4pt}

\textbf{Ablation Study for Dense Connectivity and DPM.}
We first explore the effects of dense connectivity and Dual-Path module specifically. 
A densely connected convolutional network without compression in transition layer is built as our backbone. 
The growth rate is set to 32. 
The proposed initial block and a skip architecture are adopted to construct a baseline for semantic segmentation. 
We replace the backbone with the frequently used ResNet-18\cite{he2016deep} to make a comparison between two different connection mechanisms. 
As suggested by \cite{he2019bag}, we adopt a modified ResNet, which replace the \(1\times1\) convolution with stride 2 in downsampling block with a \(2\times2\) average pooling layer with stride 2 followed by a \(1\times1\) convolution with stride 1. 
As can be seen from Table~\ref{table:table2}, our baseline is far more accurate (72.8\% VS 70.9\%) and efficient than ResNet. 
Due to feature reuse, our baseline has \(4\times\) fewer parameters and \(3\times\) fewer FLOPs than ResNet, which results in a much lower power consumption. 
To better understand the DPM, we replace the original dense layer with the proposed DPM stage by stage. 
Equipping the DPM in the last stage alone can boost the baseline by 1.8\% $(72.8\%\rightarrow74.6\%)$ in accuracy with a slight drop in inference speed. 
We continue to adopt the DPM in the third stage, which brings in 2.7\% $(72.8\%\rightarrow75.5\%)$ improvement to the baseline. 
However, adopting more DPMs at early stages does not yield further benefits. 
It verifies that feature maps from an early stage contain more spatial details, which is essential for further feature representation. 
Therefore, we adopt the original dense layer in the first two blocks and replace the original dense layer with the proposed DPM in the last two blocks as the backbone of DDPNet. 
Here, we adopt an increasing dilation rate in dense blocks to fully explore the ability to aggregate feature maps at different scales. 
For example, the dilation rate sequence of the third dense block is set to \{2, 4, 8, 16, 2, 4, 8, 16\}. 
We also try to set all dilation rates to a fixed number, or increase them gradually in different ways, which all lead to a slight decrease in accuracy. 
We compare DDPNet with PeleeNet\cite{wang2018pelee}, which utilizes a different way to aggregate multi-scale representations.
Table~\ref{table:table2} shows that DDPNet is superior to PeleeNet (75.5\% VS 72.5\%).

\setlength{\tabcolsep}{3.5pt}
\begin{table}[t]
\begin{center}
\caption{
Ablation studies for skip architecture and upsampling module. 
All models are trained from scratch on Cityscapes training set and evaluated on Cityscapes validation set.
The numbers in brackets denote the performance improvement or degradation over the baseline. 
The bold entries are the final settings for DDPNet.
}
\label{table:table3}
\begin{tabular}{|l|c|c|c|c|c|c|}
\hline
Model                         & mIoU(\%)    & Time(ms)    & FPS  & Para & FLOPs & Memory \\
\hline
Baseline                      & 69.7        & 10.7        & 93.4 & 2.4M & 12.5G & 326MB  \\
\hline
\textbf{+ Skip Architecture}  & 75.5 (+5.8) & 11.2 (+0.5) & 89.4 & 2.4M & 12.8G & 333MB  \\
\textbf{+ Upsampling Module}  & 76.2 (+6.5) & 11.7 (+1.0) & 85.4 & 2.5M & 13.2G & 412MB  \\
\hline
+ U-Net Architecture          & 75.2 (+5.5) & 11.7 (+1.0) & 85.6 & 2.5M & 13.7G & 356MB  \\
+ CARAFE\cite{wang2019carafe} & 76.2 (+6.5) & 14.8 (+4.1) & 67.4 & 2.5M & 14.0G & 601MB  \\
\hline
\end{tabular}
\end{center}
\end{table}
\setlength{\tabcolsep}{1.4pt}

\textbf{Ablation Study for Skip Architecture and Upsampling Module.}
Here, we demonstrate the effectiveness of our proposed framework for real-time semantic segmentation. 
The proposed backbone is adopted as the encoder of baseline and the output heatmap is directly upsampled to the original image size, which leads to poor segmentation of boundaries and small objects. 
We compare two different decoder structures, skip architecture and U-Net architecture, which perform skip connection on label space and feature space respectively. 
As can be seen from Table~\ref{table:table3}, skip architecture is more efficient than U-Net architecture and has a comparable accuracy (75.5\% VS 75.2\%). 
By gradually restoring spatial information, skip architecture significantly improves the baseline by 5.8\% $(69.7\% \rightarrow 75.5\%)$ in accuracy. 
Furthermore, we adopt the proposed upsampling module that leverages context information from feature maps to refine the heatmaps. 
DDPNet with the proposed framework boosts the baseline by 6.5\% $(69.7\% \rightarrow 76.2\%)$ in accuracy with a negligible drop in inference speed (1.0 ms). 
We compare another upsampling method CARAFE\cite{wang2019carafe}, which performs upsampling on feature space. 
We directly adopt CARAFE in U-Net architecture and compare it with DDPNet. 
Table~\ref{table:table3} shows that the proposed framework has a comparable accuracy, but 4 times faster (1.0 ms VS 4.1 ms). 
Some visual results and analyses are provided in the supplementary file. 

%------------------------------------------------------------------------- 
\subsection{Accuracy and Efficiency Analysis}

In this subsection, we compare the proposed DDPNet with other existing state-of-the-art real-time segmentation models on Cityscapes dataset. 
For a fair comparison, we measure the mIoU without any evaluation tricks like multi-crop, multi-scale, etc.

\setlength{\tabcolsep}{2.4pt}
\begin{table}[t]
\begin{center}
\caption{Accuracy and efficiency results on Cityscapes test dataset. 
``-'' indicates that the corresponding result is not provided by the method. 
``\(\dagger\)'' indicates that the model is evaluated on Cityscapes validation set. 
}
\label{table:table4}
\begin{tabular}{lccccccc}
\hline\noalign{\smallskip}
Method           & Pretrain & Input Size   & \#Params  & FLOPs   & GPU      & FPS   & mIoU(\%)     \\
\noalign{\smallskip}\hline\hline\noalign{\smallskip}
SegNet\cite{badrinarayanan2017segnet}   & ImageNet & 640 x 360    & 29.5M      & 286G    & TitanX M & 14.6  & 56.1     \\
ENet\cite{paszke2016enet}               & No       & 512 x 1024   & 0.4M       & 4.4G    & TitanX M & 76.9  & 58.3     \\
ESPNet\cite{mehta2018espnet}            & No       & 512 x 1024   & 0.4M       & 4.7G    & TitanX   & 112   & 60.3     \\
ERFNet\cite{romera2017erfnet}           & No       & 512 x 1024   & 2.1M       & -       & TitanX M & 41.7  & 68.0     \\
ICNet\cite{zhao2018icnet}               & ImageNet & 1024 x 2048  & 26.5M      & 29.8G   & TitanX M & 30.3  & 69.5     \\
BiSeNet1\cite{yu2018bisenet}            & ImageNet & 768 x 1536   & 5.8M       & 14.8G   & TitanX   & 72.3  & 68.4     \\
BiSeNet2\cite{yu2018bisenet}            & ImageNet & 768 x 1536   & 49.0M      & 55.3G   & TitanX   & 45.7  & 74.7     \\
DABNet\cite{li2019dabnet}               & No       & 1024 x 2048  & 0.8M       & -       & 1080Ti   & 27.7  & 70.1     \\
DFANet\cite{li2019dfanet}               & ImageNet & 1024 x 1024  & 7.8M       & 3.4G    & TitanX   & 100   & 71.3     \\
SwiftNet\(\dagger\)\cite{orsic2019defense} & No       & 1024 x 2048  & 11.8M      & 104G    & 1080Ti   & 39.9  & 70.4     \\
SwiftNet\cite{orsic2019defense}         & ImageNet & 1024 x 2048  & 11.8M      & 104G    & 1080Ti   & 39.9  & 75.5     \\
ShelfNet\cite{zhuang2019shelfnet}       & ImageNet & 1024 x 2048  & -          & -       & 1080Ti   & 36.9  & 74.8     \\
\noalign{\smallskip}\hline\hline\noalign{\smallskip}
DDPNet\(\dagger\) & No       & 768 x 1536   & 2.52M      & 13.2G   & 1080Ti   & 85.4  & 76.2    \\
DDPNet\(\dagger\) & No       & 1024 x 2048  & 2.52M      & 23.5G   & 1080Ti   & 52.6  & 77.2    \\
DDPNet         & No       & 768 x 1536   & 2.52M      & 13.2G   & 1080Ti   & 85.4  & 74.0    \\
DDPNet         & No       & 1024 x 2048  & 2.52M      & 23.5G   & 1080Ti   & 52.6  & 75.3    \\
\noalign{\smallskip}\hline
\end{tabular}
\end{center}
\end{table}
\setlength{\tabcolsep}{1.4pt}

The comparison of the accuracy (class mIoU) and efficiency (FLOPs, FPS) is shown in Table~\ref{table:table4}. 
Our DDPNet outperforms most existing real-time semantic segmentation methods in accuracy, and maintains a superior inference speed. 
Specifically, DDPNet achieves 75.3\% mIoU with 52.6 FPS for an input of \(1024\times2048\) resolution and 74.0\% mIoU with 85.4 FPS for an input of \(768\times1536\) resolution on Cityscapes test set. 
ShelfNet\cite{zhuang2019shelfnet} and SwiftNet\cite{orsic2019defense} are recent published state-of-the-art models. 
Compared to ShelfNet, DDPNet is faster (52.6 FPS VS 36.9 FPS) and more accurate (75.3\% VS 74.8\%). 
SwiftNet has a slightly better accuracy than DDPNet (75.5\% VS 75.3\%). 
However, DDPNet is trained with only Cityscapes fine-annotated images, without using any extra data. 
For a fair comparison, we compare the results of SwiftNet and DDPNet trained from scratch and evaluated on Cityscapes validation set. 
As can be seen from Table~\ref{table:table4}, DDPNet achieves significant better results in accuracy (77.2\% VS 70.4\%) and inference speed (52.6 FPS VS 39.9 FPS). 
Note that DDPNet has fewer parameters and smaller FLOPs than most methods. 
The reason is that most real-time semantic segmentation methods adopt ResNet-18 as their backbone, while DDPNet designs a light-weight and powerful backbone with fewer parameters. 

%------------------------------------------------------------------------- 
\subsection{Result on Other Dataset}

\setlength{\tabcolsep}{4.0pt}
\begin{table}[t]
\begin{center}
\caption{Accuracy results on CamVid test dataset.}
\label{table:table5}
\begin{tabular}{lccc}
\hline\noalign{\smallskip}
Method      & Pretrain   &\#Params & mIoU (\%) \\
\noalign{\smallskip}\hline\hline\noalign{\smallskip}
SegNet\cite{badrinarayanan2017segnet} & ImageNet & 29.5M   & 55.6     \\
ENet\cite{paszke2016enet}             & No       & 0.4M    & 51.3     \\
ICNet\cite{zhao2018icnet}             & ImageNet & 26.5M   & 67.1     \\
BiSeNet1\cite{yu2018bisenet}          & ImageNet & 5.8M    & 65.6     \\
BiSeNet2\cite{yu2018bisenet}          & ImageNet & 49.0M   & 68.7     \\
DABNet\cite{li2019dabnet}             & No       & 0.8M    & 66.4     \\
FPENet\cite{liu2019feature}           & No       & 0.4M    & 65.4     \\
DFANet\cite{li2019dfanet}             & ImageNet & 7.8M    & 64.7     \\
SwiftNet\cite{orsic2019defense}       & No       & 11.8M   & 63.3     \\
SwiftNet\cite{orsic2019defense}       & ImageNet & 11.8M   & 72.6     \\
FC-HarDNet\cite{chao2019hardnet}      & No       & 1.4M    & 62.9     \\
\noalign{\smallskip}\hline\hline\noalign{\smallskip}
DDPNet                              & No         & 1.1M    & 67.3     \\
DDPNet                              & Cityscapes & 1.1M    & 73.8     \\
\noalign{\smallskip}\hline
\end{tabular}
\end{center}
\end{table}
\setlength{\tabcolsep}{1.4pt}

To verify the generality of our proposed DDPNet, we conduct experiments on CamVid dataset. 
We modify the DDPNet to better fit the image resolution by
replacing the initial block with a \(3\times3\) convolution and removing the last dense block. 
As can be seen from Table~\ref{table:table5}, DDPNet achieves impressive results with only 1.1 M parameters. 
Besides, we investigate the effect of the pre-training datasets on CamVid. 
The last two rows of Table~\ref{table:table5} show that pre-training on Cityscapes can significantly improve the accuracy over 6.5\% $(67.3\% \rightarrow 73.8\%)$. 

%===========================================================
\section{Conclusion}

In this paper, we propose a novel Dense Dual-Path Network (DDPNet) for real-time semantic segmentation on high-resolution images. 
The proposed DDPNet achieves a significant better accuracy with a comparable speed and fewer parameters than most real-time semantic segmentation methods. 

%===========================================================
\section{Acknowledgement}

This work is supported by the National Key Research and Development Program of China (No.2018YFB0105103, No.2017YFA0603104, No.2018YFB0505400), the National Natural Science Foundation of China (No.U19A2069, No.U1764261, No.41801335, No.41871370), the Shanghai Science and Technology Development Foundation (No.17DZ1100202, No.16DZ1100701) and the Fundamental Research Funds for the Central Universities (No.22120180095).

%here would be your acknowledgement (if any) in the final accepted paper

%===========================================================
\bibliographystyle{splncs}
\bibliography{0188}

%this would normally be the end of your paper, but you may also have an appendix
%within the given limit of number of pages
\end{document}